\documentclass[sigplan, authorversion, screen, nonacm]{acmart}
\pdfoutput=1

\copyrightyear{2024}
\acmYear{2024}
\setcopyright{acmlicensed}\acmConference[EuroMLSys '24]{4th Workshop on Machine Learning and Systems}{April 22, 2024}{Athens, Greece}
\acmBooktitle{4th Workshop on Machine Learning and Systems (EuroMLSys '24), April 22, 2024, Athens, Greece}
\acmDOI{10.1145/3642970.3655832}
\acmISBN{979-8-4007-0541-0/24/04}

\begin{document}

\author{Pol G. Recasens}
\affiliation{
  \institution{Barcelona Supercomputing Center}
  \country{} 
}\email{pol.garcia@bsc.es}

\author{Yue Zhu}
\affiliation{
  \institution{IBM Research}
  \country{} 
}
\email{Yue.Zhu@ibm.com}

\author{Chen Wang}
\affiliation{
  \institution{IBM Research}
  \country{} 
}\email{Chen.Wang1@ibm.com}

\author{Eun Kyung Lee}
\affiliation{
  \institution{IBM Research}
  \country{} 
}\email{eunkyung.lee@us.ibm.com}

\author{Olivier Tardieu}
\affiliation{
  \institution{IBM Research}
  \country{} 
}\email{tardieu@us.ibm.com}

\author{Alaa Youssef}
\affiliation{
  \institution{IBM Research}
  \country{} 
}\email{asyousse@us.ibm.com}

\author{Jordi Torres}
\affiliation{
  \institution{Barcelona Supercomputing Center}
  \country{} 
}
\affiliation{
  \institution{Universitat Politècnica de Catalunya}
  \country{} 
}
\email{jordi.torres@bsc.es}

\author{Josep Ll. Berral}
\affiliation{
  \institution{Universitat Politècnica de Catalunya}
  \country{} 
}
\affiliation{
  \institution{Barcelona Supercomputing Center}
  \country{} 
}\email{josep.ll.berral@upc.edu}

\renewcommand{\shortauthors}{Pol G. Recasens, et al.}

\title[Towards Pareto Optimal Throughput in Small Language Model Serving]{Towards Pareto Optimal Throughput in \\ Small Language Model Serving}

\begin{abstract}
Large language models (LLMs) have revolutionized the state-of-the-art of many different natural language processing tasks. Although serving LLMs is computationally and memory demanding, the rise of Small Language Models (SLMs) offers new opportunities for resource-constrained users, who now are able to serve small models with cutting-edge performance. In this paper, we present a set of experiments designed to benchmark SLM inference at performance and energy levels. Our analysis provides a new perspective in serving, highlighting that the small memory footprint of SLMs allows for reaching the Pareto-optimal throughput within the resource capacity of a single accelerator. In this regard, we present an initial set of findings demonstrating how model replication can effectively improve resource utilization for serving SLMs.  
 
\end{abstract}

\maketitle

\section{Introduction}

Large language models (LLMs) have revolutionized the state-of-the-art of many natural language processing tasks, and show impressive zero-shot and few-shot capabilities in a wide range of applications. However, deploying language models is computationally and memory-intensive, often demanding multiple accelerators to host the model weights. For instance, the largest model from the OPT family has 175B parameters \citep{zhang2022opt}, occupying 350GB of memory space and requiring the distribution of the model across multiple devices. Although the success of LLMs was traditionally attributed to scale, recent research suggests that a curated dataset might play an important role in training high-performance models \citep{gunasekar2023textbooks, eldan2023tinystories, hoffmann2022training}. This paradigm shift, coupled with new serving optimization strategies, holds a substantial impact for a resource-constrained user, that is now able to serve SOTA small models. This rise of Small Language Models (SLMs) represents a significant step forward in making AI more accessible.

Despite the smaller size of SLMs, the incremental decoding of autoregressive language models limits the serving performance. Due to data dependencies in the self-attention layer, we process a single token per iteration, leading to matrix-vector operations. This, coupled with the large cost of loading the model weights from memory, leads to very low arithmetic intensity during single-batch inference \citep{kim2023squeezellm}. One way to increase the arithmetic intensity, defined as the ratio between arithmetic operations and bytes accessed, is to batch requests and compute multiple tokens for the same transfer of weights. As long as the memory-transfer time overlaps the compute time we can potentially improve the serving throughput by increasing the batch size. Still, after a certain size the compute time becomes non-negligible, and might grow larger than the memory transfer time. Large batches have been previously associated with compute bound scenarios \citep{kim2023squeezellm,jin2023s,aminabadi2022deepspeed}, but due to the substantial memory demands of LLMs they are rarely reached in practice. How large batches affect the serving performance of the less memory-demanding SLMs has yet to be explored.

However, batching techniques demand more memory to store key-value pairs of previously processed tokens. The space in memory dedicated to store the intermediate results of previous tokens is known as KV cache \citep{pope2023efficiently}, and handling it naively leads to memory fragmentation \citep{kwon2023efficient}. PagedAttention \citep{kwon2023efficient} algorithm identified this challenge and effectively reduced memory waste by dividing the KV cache in blocks, allowing to store KV pairs in non-contiguous memory space. In our experiments, we leverage vLLM \citep{kwon2023efficient}, a high-throughput online serving engine based on PagedAttention, to guarantee achieving the maximum batch size from our computational resources. 
 
In this paper, we provide a set of experiments to benchmark SLM inference at performance and energy levels. In this regard, we serve OPT \citep{zhang2022opt} models ranging from 125M to 13B parameters in various online scenarios, sending requests generated from the ShareGPT dataset \citep{sharegpt}. We characterise the throughput and latency trade-off when the small memory footprint allows for large batches of requests, and complement the study with internal GPU metrics, highlighting the effect of SLM inference on energy consumption. To the best of our knowledge, this provides a novel perspective, as previous inference benchmarking works are limited and primarily focused on large-scale serving \citep{samsi2023words, bai2024beyond}. From our results, we observe that the Pareto-optimal throughput with small models is reached within the resource capacity of a single accelerator. This paves the way to new optimizations, such as partitioning of GPU resources in multi-model serving. In this context, we present an initial set of findings demonstrating how model replication can improve resource utilization for serving SLMs.

\section{Background}

\subsection{KV Cache in Autoregressive Models}

The autoregressive generation of decoder-only Transformer models can be decomposed in two phases. Given an input prompt $(x_1,...,x_t)$, the model first generates the intermediate keys and values of the prompt tokens. This phase is named \textit{prefill phase}, and can be efficiently computed as each position can be processed in parallel. Then, in the \textit{autoregressive phase}, the model generates one token per iteration until completing the generation or reaching a maximum sequence length. This autoregressive behaviour is inherent to the neural attention function \citep{bahdanau2014neural}, applied through different self-attention layers in the model. This operation linearly transforms an input vector $x_t$ into query, key and value vectors $q_t, k_t, v_t $. 
\begin{equation}
    q_t = W^Q_t * x_t, k_t = W^K_t * x_t, v_t = W^V_t * x_t
\end{equation}
The output vector $o_{t}$ of the layer results from a weighted average between values from previous positions $(v_1,...,v_t)$, each weighted with its attention score $a=q_t*k^T$. Therefore, each token attends to every token in the sequence up to its position. During training, the computation of the self-attention layer can be efficiently implemented as we know the ground-truth target sequence \citep{shazeer2019fast}. However, data dependencies prevent a parallel computation during inference, resulting in the mentioned incremental computation. In this context, we are interested in storing the intermediate keys and values of each token from one iteration to the next, avoiding re-computations. The space in the GPU high-bandwidth memory (HBM) dedicated to store these intermediate results is named KV cache \citep{pope2023efficiently}, and grows with the model size and the number of tokens in the batch. Specifically, the KV cache of a token is calculated as 2 (FP16) * 2 (key and value tensors) * (hidden dimensions) * (number of layers) \citep{kwon2023efficient}. For instance, in the case of OPT 1.3B, a request with 512 input tokens and 256 output tokens requires 50 MB. Therefore, a batch of 512 requests demands at most 25.6 GB of memory space to store the KV cache.

\subsection{Batching techniques}

The low arithmetic intensity of the autoregressive phase prompts the adoption of optimization techniques to increase the resource utilization. Batching techniques improve serving performance by computing multiple requests at a time, thereby increasing the arithmetic operations per iteration, at the cost of transferring weights from the HBM to the on-chip SRAM once. It is worth noting that this also increments the amount of memory transferred due to the KV cache. \textit{Dynamic batching} involves waiting a predefined window of time to collect a group of requests sent to the engine. This approach operates at request-level granularity, leading to inefficiencies due to the waiting window and the delay in processing subsequent batch requests until all sequences in the current batch are completed. \textit{Continuous batching} addresses the problem by operating at iteration-level granularity, with the scheduler deciding at each forward pass which requests join or leave the batch. This solves the fragmentation and increases the overall throughput. However, it faces challenges, particularly in batching attention operations of requests running at different output indices. In this regard, Orca \citep{yu2022orca} proposed \textit{selective batching}, batching requests on linear operations and sequentially computing them during attention operations.

\subsection{Understanding performance limiters}
\label{sec:performance}

Performance of an inference step on a given processor can be \textit{memory-IO bound}, limited by the time spent accessing memory, or \textit{compute bound}, limited by the time spent computing operations. The metric used to measure the limiting factor is the \textit{arithmetic intensity}, defined as the ratio \textsc{ops:byte} between compute operations and bytes transferred from HBM memory. Due to the low arithmetic intensity in the autoregressive generation phase, the performance of single-batch inference is commonly classified as memory-IO bound \citep{nvidia-performance, kim2023full, kim2023squeezellm}. This scenario is frequently found in memory-limited scenarios, with large models and small batch sizes. Therefore, as long as memory-IO time overlaps compute time, we can theoretically increase the arithmetic intensity and improve the serving throughput without affecting end-to-end latency. 

However, as we increase the inner size of the matrix-matrix operations with larger batches, compute might become important. For instance, a linear layer with a large batch is usually limited by arithmetic \citep{nvidia-performance}. This might also be influenced by how continuous batching sequentially processes attention operations of different requests (see Discussion \ref{sec:discussion}). With large batches compute time grows to be larger the memory-IO time, reaching a Pareto-optimal \textit{throughput frontier}. Beyond that point, further increasing of the batch size does not improve the serving performance. The compute operations in a forward pass N parameter decoder model can be estimated $2N + 6LHQT$ add-multiply FLOPs per token seen, with L representing the number of layers, H the number of heads, Q the head dimension and T the sequence length \citep{chowdhery2023palm}.

\begin{figure}[ht]
  \centering
  \includegraphics[width=0.7\linewidth]{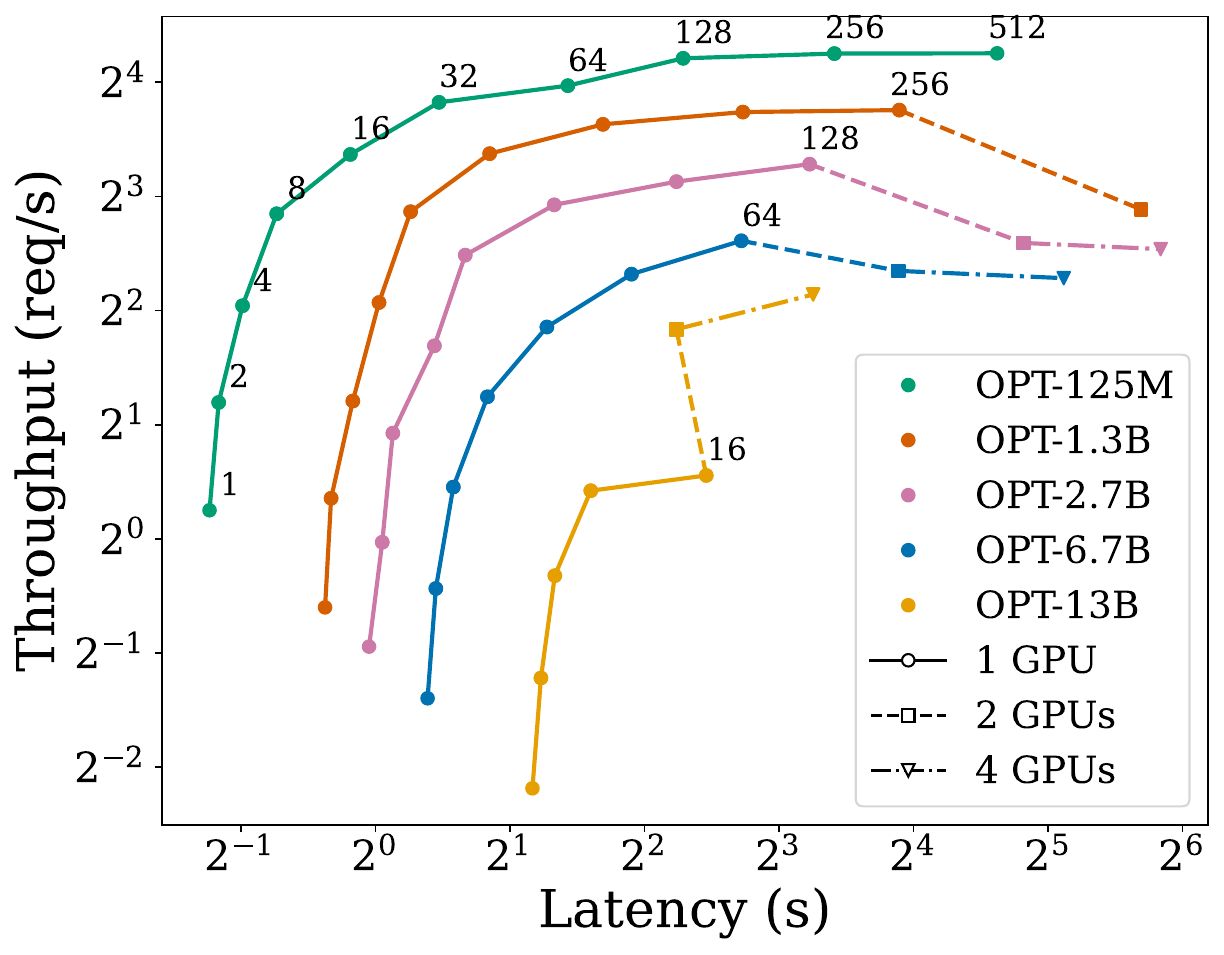}
  \caption{Throughput and latency trade-off for models of increasing size and batch sizes in powers of two. With small models we observe that throughput reaches a Pareto-optimal frontier within the resource capacity of a single accelerator.}
  \label{fig:plot_throughput}
\end{figure}

\section{Experimental setup}

In this section, we provide an overview of the hardware setup and experiment configurations employed to evaluate the inference performance of SLMs. The serving engine utilized for these experiments is vLLM\footnote{We used version v0.1.7.} \citep{kwon2023efficient}, an open-source library designed for high-throughput online serving of language models. This engine incorporates PagedAttention, a novel attention mechanism that reduces memory fragmentation in the KV cache to the minimum. This allows us to benchmark across various scenarios with the maximum possible batch size. 

The goal of our experiments is to characterise the performance of serving SLMs. Since these small models require significantly less memory than larger LLMs, we expect to batch enough requests to reach the throughput frontier within the memory available in a high-end accelerator. In our experiments, we increasingly allocate more memory to the serving system via distributed inference to find the Pareto optimality point. Other optimization techniques, such as quantization, distillation, or sparsity, can be coupled with batching. These techniques aim to reduce the memory footprint of the model, thereby creating space for larger batch sizes. We leave as future work incorporating these techniques with batching.

\subsection{Models and hardware configuration}

We employ models from the OPT \citep{zhang2022opt} family, with sizes ranging from 125M to 13B parameters. Introduced by MetaAI in May 2022, OPT belongs to decoder-only architectures such as GPT-3. It was pretrained using the self-supervised causal language modeling objective mainly over a large amount of English text. The weights of the models are provided by the Transformers library \citep{wolf2020transformers} from HuggingFace. Also, experiments on LLaMa 7B \citep{touvron2023llama}, MPT 7B \citep{MosaicML2023Introducing} and Chat-GPT J 6B \citep{gpt-j} are included for consistency. 

We use an internal IBM cluster of machines composed of 4 NVIDIA A100 GPU's  interconnected with NVLink. Each GPU has 40GB of HBM and a GPU memory bandwidth of 1555GB/s. The vLLM engine reserves 90\% of the HBM by default for parameters and the KV cache, resulting in a maximum memory allocation of 36GB for a single device. In distributed inference, vLLM leverages the Megatron-LM’s tensor parallel algorithm. This involves distributing the computation and memory usage across devices \citep{shoeybi2019megatron} using tensor and pipeline parallelism. In our context, we only use tensor parallelism, slicing and distributing the computation of the self-attention and MLP blocks to the workers.

\subsection{Workload}

We generate 500 requests from the ShareGPT dataset \citep{sharegpt}, a collection of real conversations with ChatGPT. The prompt of each request is composed by 512 input tokens, and we limit the generation to 256 output tokens. Therefore, each request requires storing KV pairs of up to 768 tokens in the KV cache, divided in blocks of 16 tokens allocated on-demand. We opt not to apply beam search so we are greedily choosing the token with highest probability, and we set \textsc{top\_p} to 1 to consider all tokens. This synthetic workload is intentionally restricted to study the effect of larger batches in a straightforward and consistent manner, providing an estimate of the amount of memory required to reach the optimal throughput. Then, the serving system can adapt to a realistic scenario with requests of varying input and output lengths. Each request is tokenized and sent to the vLLM engine through an http request, simulating a real-world deployment scenario. Although we include experiments with different arrival rates, we primarily focus on sending all the requests concurrently for various batch settings, helping to evaluate different batch configurations. It is worth noting that models larger than OPT-125m can not process the 500 requests concurrently with a single GPU.
 
\subsection{Metrics}

Batching techniques are employed to increase the serving throughput and improve resource utilization. We measure throughput, defined as the number of requests processed per second by the engine, and latency, defined as the average end-to-end processing time across all requests. Good serving performance should maximize the system throughput while providing low latency to users. In our main experimentation we consider requests of the same length, therefore the throughput and latency per token is equivalent to the throughput and latency per request normalized by the number of tokens. 

Also, we analyze the performance of the system at system's level, relying on NVIDIA internal metrics. While \textsc{nvidia-smi} is frequently used, it is not the optimal choice for performance measurements. For instance, the \textsc{gpu-utilization} metric shows the percentage of time where one or more kernels where executed on the GPU, without considering the number of streaming multiprocessors (SMs) being utilized during that time. Therefore, we use NVIDIA Data Center GPU Manager (\textsc{DCGM}) and the following metrics \citep{nvidia-dcgm}:
\begin{itemize}
    \item DCGM\_FI\_DEV\_GPU\_UTIL: Fraction of time that the GPU is not idle.
    \item DCGM\_FI\_PROF\_SM\_ACTIVE: Fraction of time that at least one warp is active on an SM, averaged across all SMs. A thread warp is considered to be active if it scheduled, regardless of whether it is actively computing or waiting for resources.
    \item DCGM\_FI\_PROF\_SM\_OCCUPANCY: Fraction of resident warps on a SM, relative to the maximum number of concurrent warps supported. 
    \item DCGM\_FI\_DEV\_POWER\_USAGE: Power usage for the device in Watts.
\end{itemize}
 



\section{Results}

\subsection{Finding the optimal batch size}

We are interested in observing how large batches affect to the inference performance. While it proves challenging to batch even a small number of requests in LLM serving, SLMs introduce a unique scenario where a single accelerator can manage the memory requirements for storing a larger number. The performance implications of this aspect have yet to be explored. In this first analysis, we benchmark the system's performance when serving small models of increasing size for different batch sizes. If the batch size cannot be achieved with a single accelerator, we distribute the serving across multiple GPUs to increase available memory. Additionally, we characterize in detail the effects of these variations on the GPU with \textsc{DCGM} metrics.

\begin{table}
  \caption{Maximum batch size in a 40 GB A100 GPU, capped at 512, for OPT models ranging in size from 125M to 13B parameters. We consider sizes in powers of two.}
  \label{tab:batchsize}
  \begin{tabular}{ccc}
    \toprule
    Model&Parameters&Maximum batch size\\
    \midrule
    OPT-125M & 250 MB & 512 \\
    OPT-1.3B & 2.6 GB & 256\\
    OPT-2.7B & 5.4 GB & 128\\
    OPT-6.7B & 13.4 GB & 64\\
    OPT-13 B & 26 GB & 16\\
  \bottomrule
\end{tabular}
\end{table}

\subsubsection{Throughput/latency analysis.}

Table \ref{tab:batchsize} shows the maximum batch size that can fit in 40GB of HBM memory from a A100 GPU for each model size, and Figure \ref{fig:plot_throughput} the throughput and latency trade-off for the different models. Each data point corresponds to a batch size ranging from 1 to the maximum attainable batch size on 4 A100 GPUs, limited at 512. In Figure \ref{fig:plot_throughput}, we notice a decline in throughput when distributing the serving to multiple GPUs, that we mainly attribute to the overhead in memory transfer across GPUs. 

In Figure \ref{fig:plot_throughput}, it is evident that with large batches we reach a throughput Pareto-optimal frontier. One observation that stands out is that for OPT 125M, OPT 1.3B and OPT 2.7B, this throughput frontier appears within the resource capacity of a single accelerator. Beyond this point doubling the size of the batch results in minimal or no improvement. For instance, in the case of OPT 125M and OPT 1.3B, after a batch size of 128 we observe a marginal improvement in the throughput at the cost of a considerable degradation in latency. However, in the case of a larger model such as OPT 13B, distributing the serving from 2 to 4 devices is still improving the throughput. This is due to the memory requirements of larger models, which scale up the KV cache size needed to reach the frontier. Following what we discussed in section \ref{sec:performance}, batching increases the arithmetic intensity and improves the serving performance. However, with large batches the compute time grows to be larger than the memory-IO time, shifting the performance limitation towards a compute-bound scenario. In the case of SLMs, this performance upper-bound opens the door to a potential slicing of the GPU resources in multi-model serving, where a scheduler assigns the optimal amount to each model depending on its individual characteristics. In that case, we have the guarantee that allocating more memory to serving a model, therefore allowing larger batch sizes, do not incur better performance. 

\begin{figure}[ht]
  \centering
  \includegraphics[width=0.8\linewidth]{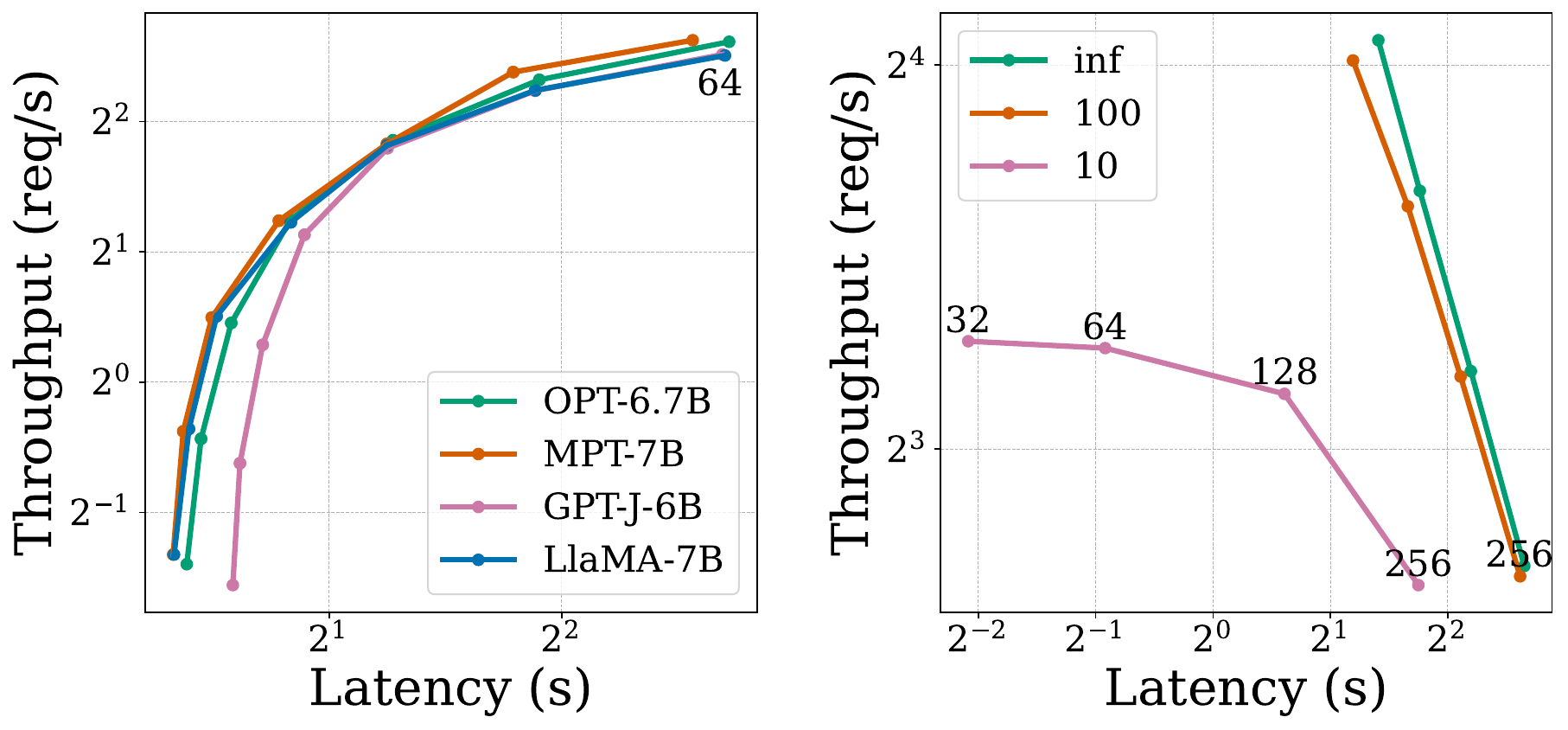}
  \caption{(Left) Throughput and latency trade-off for families of decoder models of similar size. (Right) Throughput and latency trade-off when serving OPT 6.7B with output lengths ranging from 32 to 256 in powers of two.}
  \label{fig:throughput_models}
\end{figure}

In addition, Figure \ref{fig:throughput_models} (right) shows the effect of output length in performance when serving OPT 6.7B, sending requests with different arrival rates. The waiting time for each request is sampled from a Poisson distribution. If the arrival rate is $inf$ all requests are sent instantly. We observe how increasing the number of output tokens leads to a decrease in throughput. Since the maximum amount of tokens that we can store in KV cache is defined by the hardware, increasing the output length reduces the batch size. Also, there is a slightly increase in latency. This might be related to the overhead in self-attention computation, which depends on the sentence length (see Section \ref{sec:performance}). In Figure \ref{fig:throughput_models} (left), we observe how the throughput and latency curve is consistent for different families of models. 

\subsubsection{System analysis.} \label{sec:system_analysis}
\begin{figure*}[ht]
  \centering
  \includegraphics[width=0.8\linewidth]{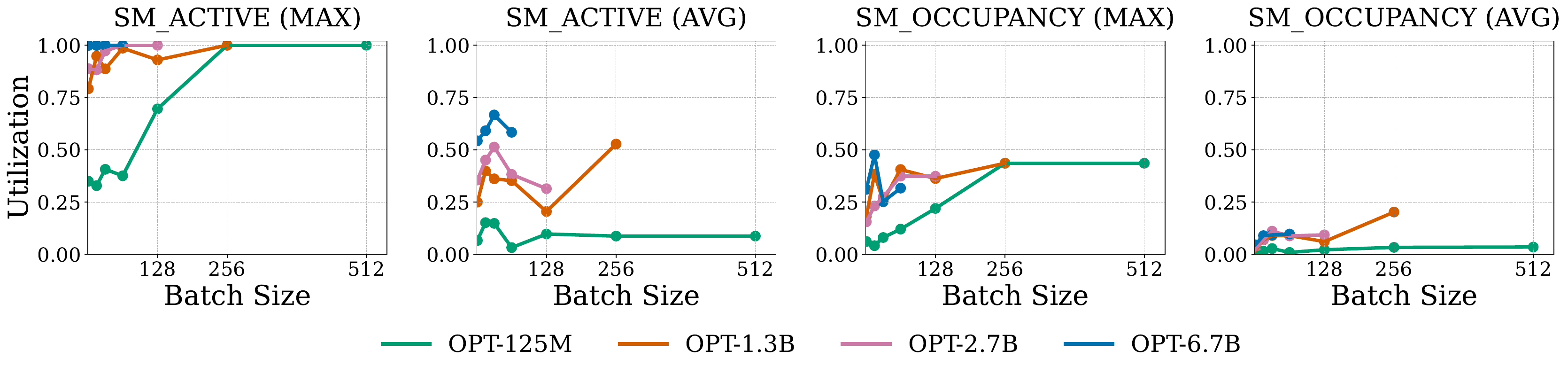}
  \caption{SM active and SM occupancy metrics when serving models ranging from 125M to 6.7B with batches in powers of two on one device. We show the mean and maximum values over all time steps, ranging from 1 to the maximum batch size.  
  }
  \label{fig:plot_gpumetrics}
\end{figure*}

Figure \ref{fig:plot_gpumetrics} shows the SM active and SM occupancy metrics for different batch sizes and models of increasing size. We show the maximum and the average values of the metrics over all cycles of the execution. The maximum SM occupancy remains low, barely reaching a 40\%, and the maximum SM activity increases with the batch size and reaches the limit. vLLM assigns one SM per sequence \citep{vllm-update} in the attention computation of the batch, leading to higher SM activity with larger batches. Although the maximum SM activity reaches the limit (at least one cycle in all the execution), the average SM activity remains low, which motivates model replication to increase the utilization of resources.

\begin{figure}[H]
  \centering
  \includegraphics[width=0.7\linewidth]{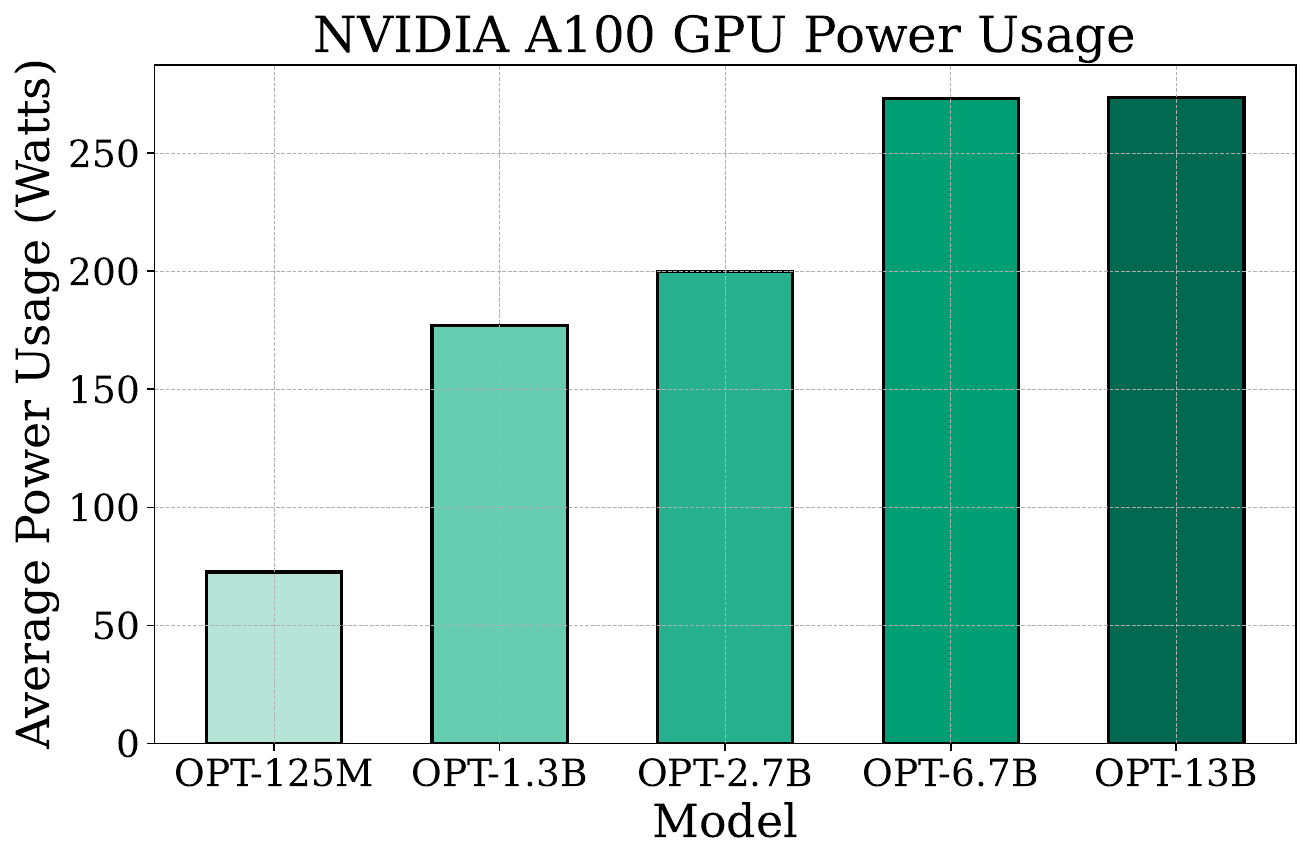}
  \caption{Average power usage in Watts of a NVIDIA A100 GPU. The value corresponds to the average power usage across all timesteps. We set the batch size of the model to the maximum in powers of two (see Table \ref{tab:batchsize}).}
  \label{fig:gpu_watts}
\end{figure}

It is also interesting to study the impact of model size on energy costs. Figure \ref{fig:gpu_watts} shows the average GPU power usage consumed by an A100 GPU when serving models of increasing size, with the maximum possible batch size (see Table \ref{tab:batchsize}). Even though we batch the maximum number of requests, the energy consumption when serving OPT 1.3B and OPT 2.7B is considerably lower than the energy consumption when serving larger models. This suggests both underutilization of resources and possible opportunities in limiting the power to reduce energy costs. Also, we observe a slight increase on energy consumption between OPT 6.7B and OPT 13B, compared to a larger increase in GPU utilization. This specific behaviour might stem from NVIDIA's thermal throttling mechanism,  which underclocks the card and restricts power consumption and heat generation.

\subsection{Model replication}

GPU resources are scarce, and our previous results show that over provisioning memory to SLMs, therefore increasing the batch size, does not necessarily correlate to a performance improvement. With this knowledge in hand, we can limit the memory allocated to each model, and run different models in the same accelerator, or replicate the same model with multiple instances. In this section, we provide a first set of findings on how we can leverage model replication to improve end-to-end serving performance. We limit the amount of memory allocated by each vLLM instance, and run multiple instances of OPT-125M, OPT-1.3B and OPT-2.7B simultaneously in the same device. Following the initial experimental setting, we generate 500 requests, but in this case they are distributed evenly across the replicas. The overall throughput is the sum of each replica's throughput, and the overall latency is the largest latency among replicas. For the sake of simplicity, we limit the batch size to the maximum attainable batch size in one device (see Table \ref{tab:batchsize}) divided by the number of replicas. 

\begin{figure}[H]
  \centering
  \includegraphics[width=1.0\linewidth]{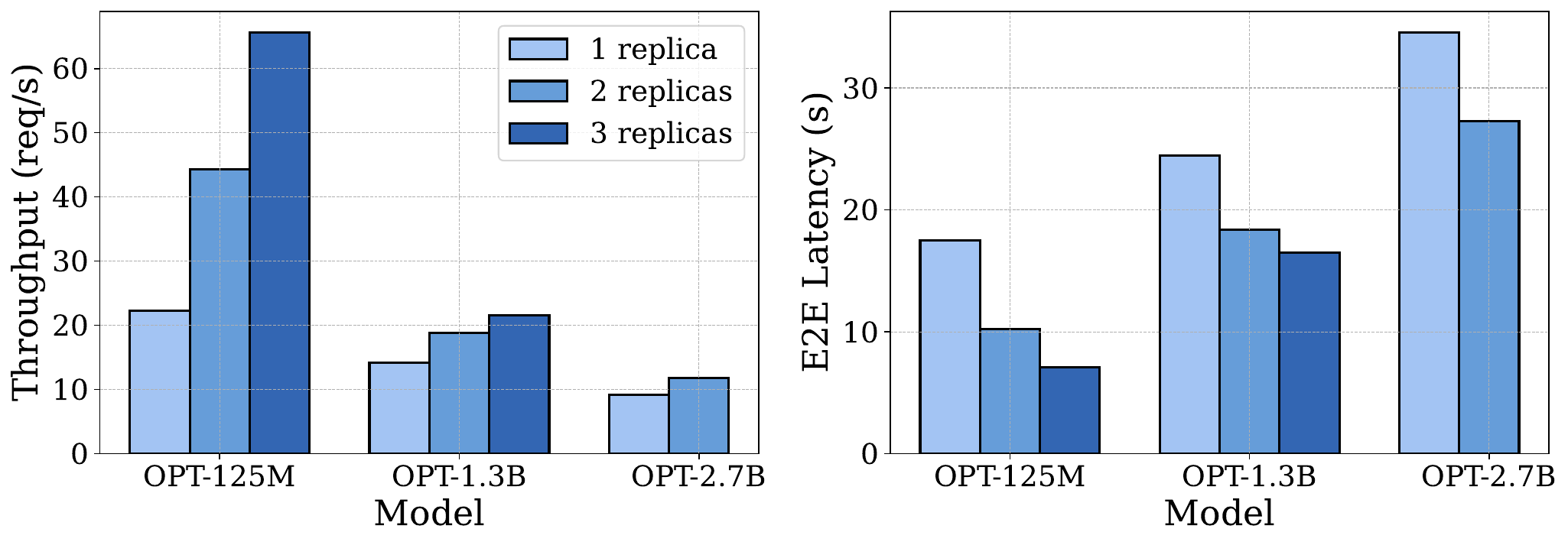}
  \caption{Throughput and latency trade-off when serving up to 3 replicas per model. We assign the same percentage of memory to each vLLM instance, all adding to 90\%.}
  \label{fig:model_replication_throughput}
\end{figure}

Following the intuition build from the results of the previous section, we partition the memory by the number of instances and observe how it affects to the overall performance. The allocated memory for each instance is independent, and the compute is shared among processes in a First Come First Serve (FCFS) manner. Figure \ref{fig:model_replication_throughput} shows an improvement in both the throughput and the latency when running multiple instances simultaneously. We observe a considerable improvement via model replication for OPT 125M, which suggest a clear underutilization at all levels of the GPU when serving really small models. Although in the case of OPT 2.7B, serving two models in the same device slightly improves the throughput, it has a positive impact on the end-to-end latency. 

\begin{figure}[ht]
  \centering
  \includegraphics[width=0.8\linewidth]{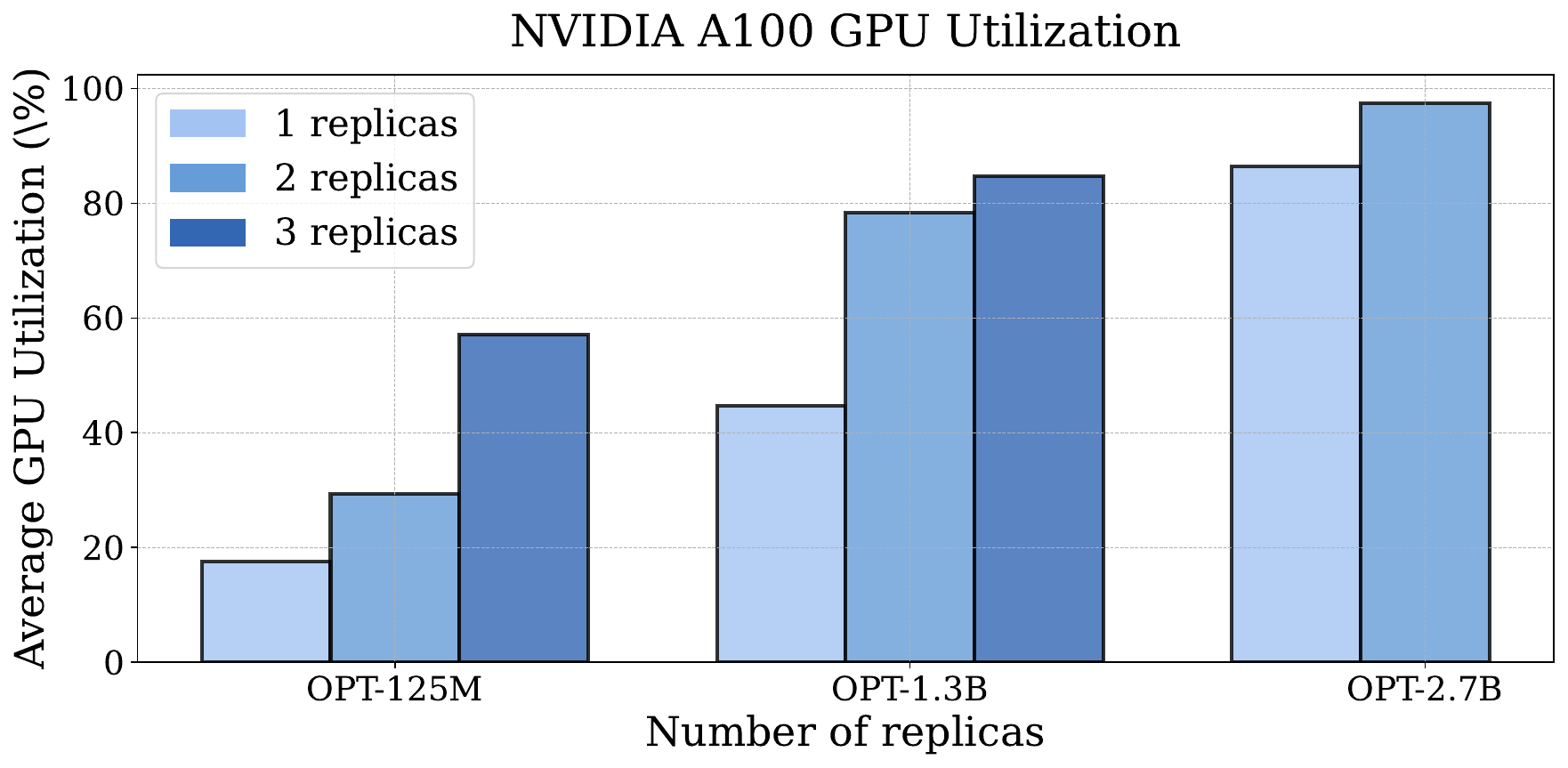}
  \caption{Average GPU utilization when serving OPT 125M, OPT 1.3B and OPT 2.7B in an NVIDIA A100 GPU. The model is replicated up to three times, if possible. }
  \label{fig:model_replication_gpu}
\end{figure}

\begin{figure}[ht]
  \centering
  \includegraphics[width=0.8\linewidth]{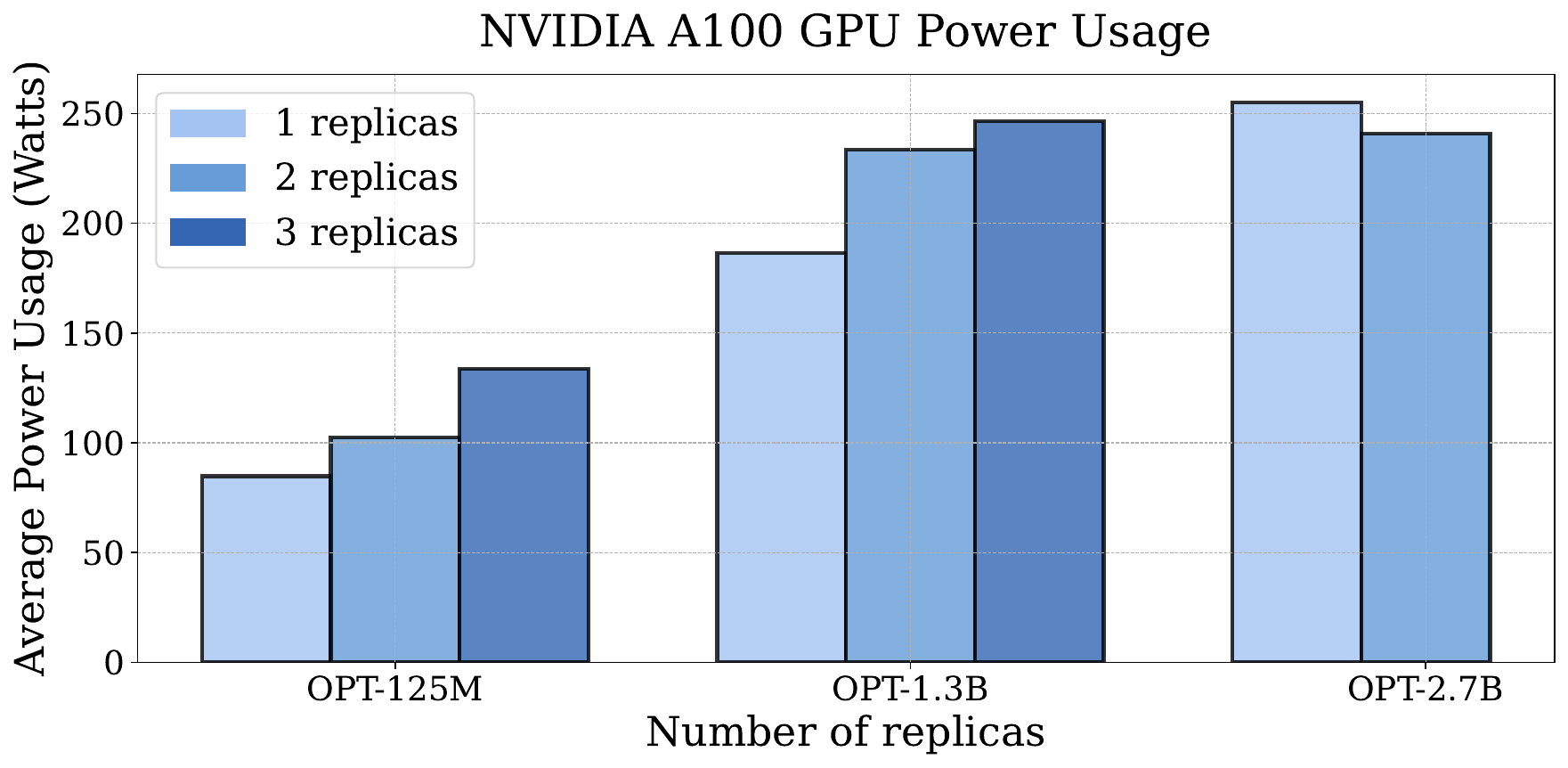}
  \caption{Average power usage in Watts when serving OPT 125M, OPT 1.3B and OPT 2.7B in an NVIDIA A100 GPU. The model is replicated up to three times, if possible. }
  \label{fig:model_replication_gpu_watts}
\end{figure}

Regarding the GPU performance, Figure \ref{fig:model_replication_gpu} shows the impact of model replication in the average GPU utilization. We observe a considerable increase in GPU utilization in the three models as we increase the number of replicas. Similarly, in Figure \ref{fig:model_replication_gpu_watts} we can observe how the average power usage increases with the number of replicas in the case of OPT 125M and OPT 1.3B, while remaining constant for OPT 2.7B. Similarly to the observation made in Section \ref{sec:system_analysis}, this might be due to the internal thermal throttling mechanism of the GPU. Therefore, we observe a positive impact of model replication on both the GPU performance and the throughput and latency of the system.

\section{Related work}

\paragraph{Serving language models.}
With LLMs already established as state-of-the-art for many different tasks, there is a growing interest in efficiently deploying and serving language models. In this regard, general online serving systems have emerged, such Orca \citep{yu2022orca}, Text Generation Inference (TGI) \citep{hf-tgi}, DeepSpeed-FastGen \citep{aminabadi2022deepspeed,deepspeed-fastgen} or vLLM \citep{kwon2023efficient}, which improve serving performance through different optimizations. Other serving systems, like AlpaServe \citep{li2023alpaserve}, optimize multi-model serving via statistical multiplexing of multiple devices. Additionally, engines such as FasterTransformer \citep{nvidia-fastertransformer} provide a highly optimized implementation of language models written in C++/CUDA.

\paragraph{Memory optimizations.} 
Memory size is lagging behind the increasing compute capabilities in modern accelerators. Therefore, several approaches aim to reduce the resource requirements for serving language models. In this regard, quantization \citep{xiao2023smoothquant,frantar2022gptq, sheng2023flexgen} minimizes the memory footprint by compressing the model weights, sparsity \citep{wang2021spatten} prunes unimportant tokens and heads, and offloading techniques such as \citep{sheng2023flexgen} leverage memory from CPU and disk in offline serving scenarios. We leave as further work benchmarking the systems performance while coupling batching with these optimizations, that indirectly increase the effective batch size.

\paragraph{Managing the KV cache.}
The management of the KV cache in LLM serving is challenging due to the unpredictable number of tokens generated per request, and it limits the number of requests that can be effectively batched. Orca's \citep{yu2022orca} initial approach pre-allocates contiguous GPU memory for the maximum possible output length, introducing memory fragmentations. Recent works identified and addressed this problem. S3 \citep{jin2023s} proposes to predict the number of output tokens of each request, adjusting the amount of memory that is pre-allocated. On the other hand, \textsc{vLLM} \citep{kwon2023efficient} introduces PagedAttention, an attention algorithm inspired on OS paging that divides the KV cache in blocks of tokens, and stores them in non-contiguous memory blocks. This approach dynamically allocates blocks on-demand, and allows memory sharing in complex decoding scenarios (e.g parallel sampling). The KV cache can be further optimized to reuse attention states across different requests. In this regard, the prompt cache \citep{gim2023prompt} precomputes attention states of frequently visited text segments and reuses them for different sequences. Also, RadixAttention mantains a radix tree on the CPU to reuse the KV cache during runtime. For our experiments, we adhere to vLLM and PagedAttention to control the KV cache size for each request. 

\section{Discussion}
\label{sec:discussion}
\begin{figure}[ht]
  \centering
  \includegraphics[width=0.7\linewidth]{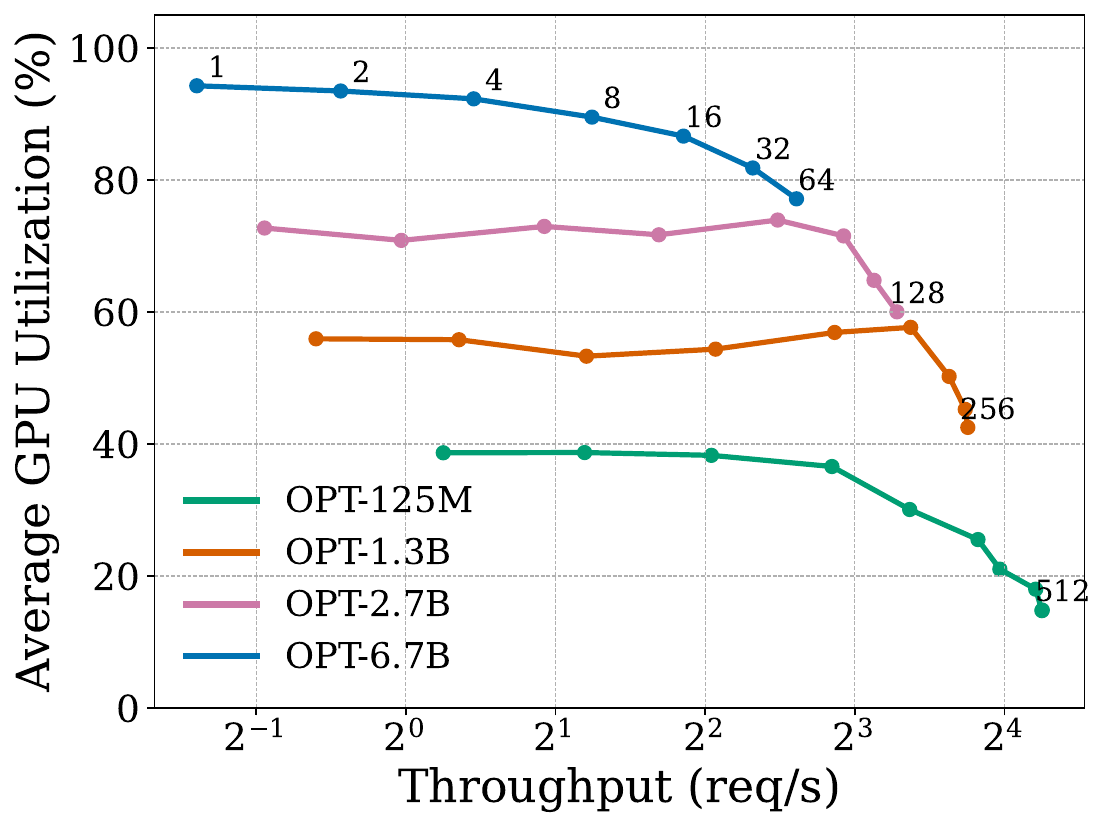}
  \caption{Average GPU utilization and throughput for each model size and batch size.}
  \label{fig:gpu_util}
\end{figure}

The growing interest in the topic of LLM inference leads to never-ending improvements in serving systems. Therefore, our experiments are linked to the performance of the current implementation of vLLM. Although we do not couple batching with other optimization techniques to mitigate potential issues, there are some inherent limitations in the implementation. For instance, PyTorch networks that are running on high-throughput GPUs are frequently limited by CPU overheads \citep{cudagraph}, that might take up to a 50\% of the latency \citep{vllm-update}. This leads to unusual behaviour in GPU utilization metric: while increasing the batch size in increases the overall throughput (see Figure \ref{fig:gpu_util}), the GPU utilization remains constant, and decreases with large batches. However, this behaviour is orthogonal to the goal of the paper, where we aim to characterize the Pareto-optimal throughput for small models. Further versions of vLLM introduce CUDA Graphs, launching multiple GPU operations through a single CPU operation, therefore reducing the launching overheads \citep{cudagraph}. Although we could have used a serving system written in optimized C++ with less overheads, such as FasterTransformer \citep{nvidia-fastertransformer}, we would sacrifice PagedAttention. In this regard, we leave as future work analyzing at kernel level the performance of the system to investigate further in this limitation. 

Moreover, our baseline device is a 40GB NVIDIA A100 GPU, which may not be affordable for the average user. While this device offers enough memory to reach a throughput limit for small models, cheaper devices might not. Given the optimal amount of memory that we need to allocate to reach the throughput frontier, we must consider a cost trade-off between serving SLMs on smaller devices or replicate the model on larger devices. It is also worth noting that in our experiments we set the input length of each request to be 512 tokens, and we limit the output length to 256. Although this restricted configuration allows us to characterize the Pareto optimal throughput, in more realistic scenarios requests might vary in input and output length. We leave as a future analysis the benchmark of SLM serving with heterogeneous devices, and with heterogeneous requests. Also, the study of more suitable alternatives for model replication such as Multi-Process Service (MPS) or Multi-Instance GPU (MIG), providing multiple users with separate GPU resources for optimal GPU utilization. 

\section{Conclusion}

This paper characterizes the serving performance of SLMs, highlighting the implications of memory allocation on inference throughput. Our analysis shows that for small models a single high-end accelerator has enough memory to reach a Pareto-optimal throughput frontier given a large batch of requests. Beyond that point, allocating more memory results in minimal or no improvements. In light of our results, we pave the way for new optimizations in model serving, presenting an initial set of findings that show how model replication on a single device improves overall inference performance. Further analysis should consider a more realistic serving scenario with heterogeneous requests and devices, and explore model replication with more suitable techniques.


\section*{Acknowledgments}

This work has been partially financed by grant agreement EU-HORIZON GA.101095717 and by the EU-HORIZON MSCA programme under grant agreement EU-HORIZON MSCA GA.101086248. Also, it has been partially financed by Generalitat de Catalunya (AGAUR) under grant agreement 2021-SGR-00478, and by the Spanish Ministry of Science (MICINN), the Research State Agency (AEI) and European Regional Development Funds (ERDF/FEDER) under grant agreement PID2021-126248OB-I00, MCIN/AEI/
10.13039/ 501100011033/ FEDER, UE.

\bibliographystyle{ACM-Reference-Format}
\bibliography{arxiv/ms}

\appendix

\end{document}